%% file: ms.tex
\documentclass[10pt,twocolumn,letterpaper]{article}

\usepackage{iccv}
\usepackage{times}
\usepackage{epsfig}
\usepackage{graphicx}
\usepackage{amsmath}
\usepackage{booktabs}
\usepackage{enumitem}
\usepackage{amssymb}
\usepackage{authblk}
\usepackage{numprint}
\usepackage{caption}
\usepackage{graphicx}
\usepackage{subfig}
\usepackage{tabularx}
\usepackage{siunitx}
\usepackage{float}
\usepackage{amsfonts}
\usepackage{dblfloatfix}    
\usepackage{comment}
\usepackage[printonlyused]{acronym}
\usepackage[pagebackref=true,breaklinks=true,letterpaper=true,colorlinks,bookmarks=false]{hyperref}

\iccvfinalcopy 

\def\iccvPaperID{862} 

\begin{document}

\title{GLAMpoints: Greedily Learned Accurate Match points}

\author{Prune Truong$^{1,2}$\thanks{Work produced during an internship at RetinAI Medical AG} \quad Stefanos Apostolopoulos$^1$ \quad Agata Mosinska$^1$ \quad Samuel Stucky$^1$ \quad Carlos Ciller$^1$ \quad Sandro~De~Zanet$^1$ \\
$^1$RetinAI Medical AG, Switzerland \quad $^2$ETH Zurich, Switzerland \\
{\tt\small prune.truong@vision.ee.ethz.ch \quad \{stefanos, agata, samuel, carlos, sandro\}@retinai.com}
}

\maketitle
\input{acronyms.tex}

\begin{abstract}
    \input{abstract}

\end{abstract}

\input{01_introduction.tex}

\input{02_State_of_the_Art.tex}

\input{03_Method.tex}

\input{04_Experimental_Validation.tex}

\input{05_Conclusion.tex}

{\small
\bibliographystyle{ieee_fullname}
\bibliography{string,RetinAI}
}

\appendix

\begin{table*}[!b]
\centering
\caption{Evaluation metrics calculated over 206 pre-processed pairs of the \textit{slitlamp} dataset. }\label{tab:SIFT_metrics}\vspace{-2mm}
\resizebox{0.82\textwidth}{!}{%
\begin{tabular}{@{}lll@{}}
\toprule
                                             & SIFT with rotation invariance & SIFT rotation-dependent \\ \midrule
Success rate of Acceptable Registrations [\%] & 49.03                         & \textbf{50.49}                            \\
Success rate of Inaccurate Registrations [\%] & 50.49                         & 47.57                           \\
Success rate of Failed Registrations [\%]     & 0.49                         & 1.94                             \\
M.score                                      & 0.0470           & \textbf{0.056}                \\
Coverage Fraction                            & 0.1348            & \textbf{0.15}                 \\
AUC                                          & 0.1274            & \textbf{0.143}                \\ \bottomrule
\end{tabular}%
}\vspace{2mm}
\end{table*}

\begin{table*}[b!]
\centering
\caption{Parameter range used for random homography generation during training.}\label{tab:homo_generation}\vspace{-2mm}
\resizebox{\textwidth}{!}{%
\begin{tabular}{@{}llllllllll@{}}
\toprule
\multicolumn{2}{c}{\textbf{Scaling}} & \multicolumn{2}{c}{\textbf{Perspective}} & \multicolumn{2}{c}{\textbf{Translation}} & \multicolumn{2}{c}{\textbf{Shearing}}           & \multicolumn{1}{c}{\textbf{Rotation}} & \multicolumn{1}{c}{\textbf{}} \\
\midrule
min scaling           & 0.7          & min perspective parameter   & 0.000001   & max horizontal displacement     & 100    & min/max horizontal shearing & -0.2 / 0.2          & max angle                             & 25                            \\
max scaling           & 1.3          & max perspective parameter   & 0.0008     & max vertical displacement       & 100    & min/max        vertical shearing &           -0.2 / 0.2                            &      \\       
\bottomrule
\end{tabular}%

}
\end{table*} 

\newpage
\input{06_Supplementary_Material.tex}

\end{document}

%% file: acronyms.tex
\acrodef{CNN}{Convolutional Neural Network}
\acrodef{ROC}{Receiving Operating Characteristic}
\acrodef{AMD}{Age-related Macular Degeneration}
\acrodef{DoG}{Difference Of Gaussian}
\acrodef{NMS}{Non-Max-Supression}
\acrodef{SIFT}{Scale-Invariant Feature Transform}
\acrodef{RanSaC}{Random Sampling Consensus}
\acrodef{RL}{Reinforcement Learning}
\acrodef{NNDR}{Nearest Neighbor Distance Ratio}
\acrodef{DR}{Diabetic Retinopathy}
\acrodef{SURF}{Speeded-Up Robust Features}
\acrodef{FIRE}{Fundus Image Registration}
\acrodef{ORB}{Oriented Fast and Rotated Brief}
\acrodef{FREAK}{Fast Retina Keypoint}
\acrodef{BRISK}{Binary Robust Invariant Scalable Keypoints}
\acrodef{SfM}{Structure from Motion}
\acrodef{LIFT}{Learned Invariant Feature Transform  }
\acrodef{GLAMpoints}{Greedily Learned Accurate Match Points}
\acrodef{SLAM}{Simultaneous Localization and Mapping}
\acrodef{ReLU}{Rectified Linear Unit}
\acrodef{UNet}{U-Net}
\acrodef{LF-NET}{Local Feature Network}

%% file: abstract.tex
We introduce a novel CNN-based feature point detector - \ac{GLAMpoints} - learned in a semi-supervised manner. Our detector extracts repeatable, stable interest points with a dense coverage, specifically designed to maximize the correct matching in a specific domain, which is in contrast to conventional techniques that optimize indirect metrics. In this paper, we apply our method on challenging retinal slitlamp images, for which classical detectors yield unsatisfactory results due to low image quality and insufficient amount of low-level features. We show that GLAMpoints significantly outperforms classical detectors as well as state-of-the-art CNN-based methods in matching and registration quality for retinal images. Our method can also be extended to other domains, such as natural images. Training code and model weights are available at \url{https://github.com/PruneTruong/GLAMpoints_pytorch}.

%% file: 01_introduction.tex
\section{Introduction}

Digital fundus images of the human retina are widely used to diagnose variety of eye diseases, such as \ac{DR}, glaucoma, and \ac{AMD} \cite{Sanchez-galeana2001, Zhou1994}. For retinal images acquired during the same session and presenting small overlaps, image registration can be used to create mosaics depicting larger areas of the retina. Through image mosaicking, ophthalmologists can display the retina in one large picture, which is helpful during diagnosis and treatment planning. Besides, mosaicking of retinal images taken at different time points has been shown to be important for monitoring the progression or identification of eye diseases. More importantly, fundus image registration has been explored in eye laser treatment for \ac{DR}. It allows real-time tracking of the vessels during surgical operations to ensure accurate application of the laser on the retina and minimal damage to the healthy tissues.  

\begin{figure}
\centering
\vspace{-2mm}\begin{tabular}{c c}
\hspace{-2.5mm}\includegraphics[width=0.23\textwidth]{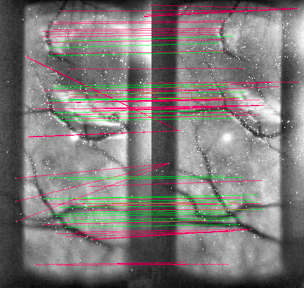} & 
\hspace{-1.5mm}\includegraphics[width=0.23\textwidth]{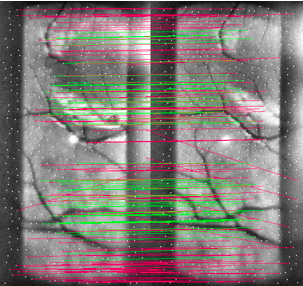} \\
\hspace{-2.5mm}\includegraphics[width=0.23\textwidth]{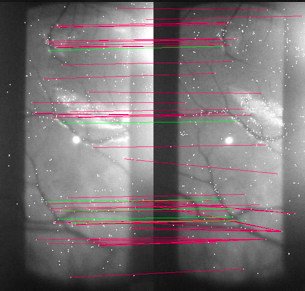} & 
\hspace{-1.5mm}\includegraphics[width=0.23\textwidth]{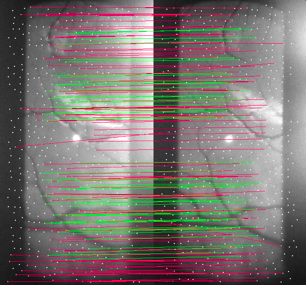} \\
a) SIFT & b) GLAMpoints
\end{tabular}
\vspace{-3mm}
\caption{Keypoints detected by SIFT and GLAMpoints and resulting matches for a pair of pre-processed (top) and raw (bottom) slitlamp images. Detected points are in white, green matches are true positive, red represents false positive. Our GLAMpoints detector produces more reliable keypoints even without additional pre-processing.}
\label{matches-slitlamp}
\vspace{-5mm}
\end{figure}

Mosaicking usually relies on extracting repeatable interest points from the images, matching the correspondences and searching for transformations relating them. As a result, the keypoint detection is the first and the most crucial stage of this pipeline, as it conditions all further steps and therefore the success of the registration.

At the same time, classical feature detectors are general-purpose and manually optimized for outdoor, in-focus, low-noise images with sharp edges and corners. They usually fail to work with medical images, which can be distorted, noisy, have no guarantee of focus and depict soft tissue with no sharp edges (see Figure~\ref{fig:ex_dataset}). Traditional methods perform sub-optimally on such images, making more sophisticated optimization necessary at a later step in the registration, such as \ac{RanSaC}~\cite{RANSAC}, bundle adjustment~\cite{bundle} and \ac{SLAM}~\cite{slam} techniques. Besides, supervised learning methods for keypoint detection fail or are not applicable, due to missing ground truths for feature points.

In this paper we present a method for learning feature points in a semi-supervised manner. Learned feature detectors were shown to outperform the heuristics-based methods, but they are usually optimized for repeatability, which is a proxy for the matching quality and as a result they may underperform during the final matching. On the contrary, our keypoints - \ac{GLAMpoints} - are trained for the final matching accuracy and when associated with \ac{SIFT}~\cite{sift} descriptor they outperform state-of-the-art in matching performance and registration quality on retinal images. As shown in Figure~\ref{matches-slitlamp}, \ac{GLAMpoints} produces significantly more correct matches than \ac{SIFT} detector. 

Registration based on feature points is inherently non-differentiable due to point matching and transformation estimations. We take inspiration from the loss formulation in \ac{RL} using a reward to compute the suitability of the detected keypoints based on the final registration quality. It makes it possible to use the key performance measure, \ie matching power, to directly train a \ac{CNN}. Our contribution is therefore a formulation for keypoint detection that is directly optimized for the final matching performance in an image domain. Both training code and model weights are available at \cite{glampoint_tensorflow} for the Tensorflow version and \cite{glampoint_pytorch} for the PyTorch variant.

The remainder of this paper is organized as follows: we introduce the current state-of-the-art feature detection methods in section \ref{sec:related_work}, our training procedure and loss in section \ref{sec:method}, followed by experimental comparison of previous methods in section \ref{sec:results} and conclusion in section \ref{sec:conclusion}.

%% file: 02_State_of_the_Art.tex
\section{Related Work}
\label{sec:related_work}

Existing registration algorithms can be classified as area-based and feature-based approaches. The former typically rely on a similarity metric such as cross-correlation~\cite{Cideciyan1995}, mutual information~\cite{Pluim2003,Legg2013} or phase correlation~\cite{Huang2005} to compare the intensity patterns of an image pair and estimate the transformation. However, in the case of changes in illumination or small overlapping areas, the application of area-based approaches becomes challenging or infeasible. Conversely, feature-based methods extract corresponding points on pairs of images along with a set of features and search for a transformation that minimizes the distance between the detected key points. Compared with area-based registration techniques, they are more robust to changes of intensity, scale and rotation and therefore, they are considered more appropriate for
problems such as medical image registration.

Typically, feature extraction and matching of two images comprise four steps: detection of interest points, computing feature descriptor for each of them, matching of corresponding keypoints and estimation of a transformation between the images using the matches. As can be seen, the detection step influences every further step and is therefore crucial for a successful registration. It requires a high image coverage and stable key points in low contrasts images.

In the literature, local interest point detectors have been thoroughly studied. \ac{SIFT}~\cite{Lowe2004} is probably the most well known detector/descriptor in computer vision. It computes corners and blobs on different scales to achieve scale invariance and extracts descriptors using the local gradients. \ac{SURF}~\cite{Bay2006} is a faster alternative, using Haar filters and integral images, while KAZE~\cite{Alcantarilla2012} exploits non-linear scale space for more accurate keypoint detection.

In the field of fundus imaging, a widely used technique relies on vascular trees and branch point analysis~\cite{Li2017,Hang}. However, accurate segmentation of the vascular trees is challenging and registration often fails on images with few vessels.  Alternative registration techniques are based on matching repeatable local features; Chen~\etal ~\cite{Chen2010} detected Harris corners~\cite{harris} on low quality multi-modal retinal images and assigned them a partial intensity invariant feature (Harris-PIIFD) descriptor. They achieved good results on low quality images with an overlapping area greater than 30\%, but the method is characterised by low repeatability. Wang~\etal~\cite{Wang2015} used \ac{SURF} features to increase the repeatability and introduced a new method for point matching to reject a large number of outliers, but the success rate drops significantly when the overlapping area diminishes below 50\%. Cattin~\etal~\cite{Cattin} also demonstrated that \ac{SURF} can be efficiently used to create mosaics of retina images even for cases with no discernible vascularisation. However this technique only appeared successful in the case of highly self-similar images. D-saddle detector/descriptor~\cite{Ramli} was shown to outperform the previous methods in terms of rate of successful registration on the \ac{FIRE} Dataset~\cite{Hernandez-matas2017},  enabling the detection of interest points on low quality regions. 

Recently, with the advent of deep learning, learned detectors based on \ac{CNN} architectures were shown to outperform state-of-the-art computer vision detectors~\cite{Fisher,Detone,LIFT,Trulls,Johns}. 
\ac{LIFT}~\cite{LIFT} uses patches to train a fully differentiable deep \ac{CNN} for interest point detection, orientation estimation and descriptor computation based on supervision from classical \ac{SfM} systems. SuperPoint~\cite{Detone} introduced a self-supervised framework for training interest point detectors and descriptors. It rises to state-of-the-art homography estimation results on HPatches~\cite{Lenc} when compared to \ac{SIFT}, \ac{LIFT} and \ac{ORB}~\cite{Rublee2011}. The training procedure is, however, complicated and their self-supervision implies that the network can only find corner points. Altwaijry \etal~\cite{Altwaijry16} proposed a two-step CNN for matching aerial image patches, which is a particularly challenging task due to ultra-wide baseline. Altwaijry \etal~\cite{Altwaijry2016b} also introduced a method to detect keypoint locations on different scales, utilizing high activations in recursive network feature maps. KCNN~\cite{Febbo18} was shown to emulate hand-crafted detectors by training small networks using keypoints detected by other methods as ground-truth.
\ac{LF-NET}~\cite{Trulls} is the closest to our method: a keypoint detector and descriptor is trained end-to-end  in a two branch set-up, one being differentiable and feeding on the output of the other non-differentiable branch. However, they optimized their detector for repeatability between image pairs, not taking into account the matching performance.

Truong~\etal~\cite{truong} presented an evaluation of \ac{SURF}, KAZE, \ac{ORB}, \ac{BRISK}~\cite{Leutenegger2011}, \ac{FREAK}~\cite{Alahi2012}, \ac{LIFT}, SuperPoint and LF-NET both in terms of image matching and registration quality on retinal fundus images. They found that while SuperPoint outperforms all the others relative to the matching performance, \ac{LIFT} demonstrates the highest results in terms of registration quality, closely followed by KAZE and \ac{SIFT}. The highlighted issue was that even the best-performing detectors produce feature points which are densely positioned and as a result may be associated with a similar descriptor. This can lead to false matches and thus inaccurate or failed registrations. 

Our goal is to tackle this problem by introducing a novel semi-supervised learned method for keypoint detection. Detectors are often optimized for repeatability (such as \ac{LF-NET}~\cite{Trulls}) and not for the quality of the associated matches between image pairs. Our training procedure uses a reward concept akin to \ac{RL} to extract repeatable, stable interest points with a uniform coverage and it is specifically designed to maximize correct matching on a specific domain, as shown for challenging retinal slit lamp images.

%% file: 03_Method.tex
\section{Methods}
\label{sec:method}

\begin{figure*}
\begin{center}
\includegraphics[width=0.99\linewidth]{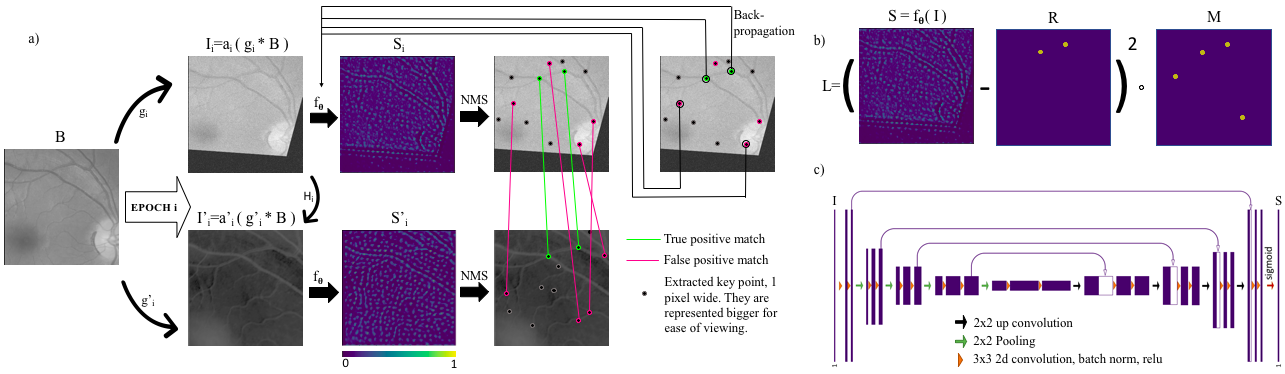}
\end{center}
\vspace{-1mm}\caption{a) Training steps for an image pair $I_i$ and $I'_i$ at epoch $i$ created from a particular base image $B$. $I_i$ and $I'_i$ are created by warping B according to homographies $g_i$ and $g'_i$ respectively. $a_i$ and $a'_i$ refer to the additional appearance augmentations applied to each image. b) Loss computation corresponding to situation a. c) Schematic representation of Unet-4.}
\label{fig:method}\vspace{-2mm}
\end{figure*}

Our trained network predicts the location of stable interest points, called \ac{GLAMpoints}, on a full-sized gray-scale image. 
In this section, we explain how our training set was produced and our training procedure. As we used standard convolutional network architecture, we only briefly  discuss it in the end.

\subsection{Dataset}
We trained our model on a dataset from the ophthalmic field, namely slit lamp fundus videos, used in laser treatment (examples in Figure~\ref{fig:ex_dataset}). In this application, live registration is required for an accurate ablation of the retinal tissue. Our training dataset consists of 1336 images with different resolutions, ranging from \SIrange{300}{700}{px} by \SIrange{150}{400}{px}. These images were acquired with multiple cameras and devices to cover large variability of appearances. They come from eye examination of 10 different patients, who were healthy or with diabetic retinopathy. 

From the original fundus images, image pairs are synthetically created and used for training. Let B be a particular base image from the training dataset, of size $H \times W$.
At every step $i$, an image pair $I_{i}, I'_{i}$ is generated from image $B$ by applying two separate, randomly sampled homography transforms $g_{i}, g'_{i}$.  Images $I_i$ and $I'_i$ are thus related according to the homography $H_{I_i,I'_i}=g_i'*g_i^{-1}$ (see supplementary material). 
On top of the geometric transformations, standard data augmentation methods are used: gaussian noise, changes of contrast, illumination, gamma, motion blur and the inverse of image. A subset of these appearance transformations is randomly chosen for each image of the pair. 

\subsection{Training}

We define our learned function $f_\theta (I) \xrightarrow{} S$, where $S$ denotes the pixel-wise feature point probability map of size $H \times W$.
Lacking a direct ground truth of keypoint locations, a delayed reward can be  computed instead. We base this reward on the matching success, computed \textit{after registration}. The training proceeds as follows:
\begin{enumerate}
\item Given a pair of images $I \in \mathbb{R}^{H\times W}$ and $I' \in \mathbb{R}^{H\times W}$ related with the ground truth homography $H=H_{I, I'}$, our model provides a score map for each image, $S=f_\theta(I)$ and $S'=f_\theta(I')$. 

\item The locations of interest points are extracted on both score maps using standard non-differentiable \ac{NMS}, with a window size $w$.

\item A 128 root-\ac{SIFT}~\cite{Arandjelovi} feature descriptor is computed for each detected keypoint. 

\item The keypoints from image $I$ are matched to those of image $I'$ and vice versa using a brute force matcher \cite{brute}. Only the matches that are found in both directions are kept. 

\item The matches are checked according to the ground truth homography $H$. A match is defined as true positive if the corresponding keypoint $x$ in image $I$ falls into an $\epsilon$-neighborhood of the point $x'$ in $I'$ after applying $H$. This is formulated as $\left \|  H*x-x'  \right \| \leq \varepsilon$, where we chose  $\varepsilon$ as \SI{3}{px}. 
\end{enumerate} 
Let $T$ denote the set of true positive key points. If a given detected feature point ends up in the set of true positive points, it gets a positive reward. All other points/pixels are given a reward of 0. Consequently, the reward matrix $R \in \mathbb{R}^{H\times W}$ for a keypoint $(x,y)$ can be defined as follows: 

\begin{equation}
    R_{x,y}  = \left\{\begin{array}{@{}rl@{}}
        1, & \mbox{for}\quad {(x,y)} \in T\\
        0, & \mbox{otherwise}
        
        \end{array}\right\}
\end{equation}

This leads to the following loss function:
\begin{equation}
   L_{\mbox{simple}}(\theta,I) = \sum (f_\theta(I) - R)^2 
\end{equation}
However, a major drawback of this formulation is the large class imbalance between positively rewarded points and null-rewarded ones, where latter prevails by far, especially in the first stages of training. Given a reward $R$ with mostly zero values, the $f_\theta$ converges to a zero output. 
Hard mining has been shown to boost training of descriptors~\cite{Simo-serra15}. Thus, negative hard mining on the false positive matches might also enhance performance in our method, but has not been investigated in this work. 

Instead, to counteract the imbalance, we use \emph{sample mining}: we select all $n$ true positive points and randomly sample additional $n$ from the set of false positives. We only back-propagate through the $2n$ true positive feature points and mined false positive key points. If there are more true positives than false positives, gradients are backpropagated through all found matches. This mining is mathematically formulated as a binary pixel-wise mask $M$, equal to $1$ at the locations of the true positive key points and that of the subset of mined feature points, and equal to 0 otherwise.
The final loss is thus formulated as follows: 
\begin{equation}
   L(\theta,I)=\frac{\sum (f_{\theta}(I)-R)^{2} \cdot M }{\sum\raisebox{-0.75pt}{$M$}}
\end{equation}where $\cdot$ denotes the element-wise multiplication. 
\\

An overview of the training steps is given in Figure~\ref{fig:method}. Importantly, only step 1 is differentiable with respect to the loss. We learn directly on a reward which is the result of non differentiable actions, without supervision. 

It should be noted that the descriptor we used is the root-SIFT version without rotation invariance. The reason is that it performs better on slitlamp images than root-\ac{SIFT} detector/descriptor with rotation invariance  (see supplementary material for details). The aim of this paper is to investigate the detector only and therefore we used rotation-dependent root-\ac{SIFT} for consistency.

\subsection{Network}

A standard 4-level deep Unet~\cite{ronneberger} with a final sigmoid activation was used to learn $f_\theta$. It comprises of 3x3 convolution blocks with batch normalization and \ac{ReLU} activations (see Figure~\ref{fig:method},c). Since the task of keypoint detection is similar to pixel-wise binary segmentation (class interest point or not), Unet was a promising choice due to its past successes in binary and semantic segmentation tasks.

%% file: 04_Experimental_Validation.tex
\section{Results}
\label{sec:results}

In this section, we describe the testing dataset and the evaluation protocol. We then compare state-of-the-art detectors, quantitatively and qualitatively to our proposed \ac{GLAMpoints}.

\subsection{Testing datasets}

\begin{figure}[t]
\centering
\includegraphics[width=0.45\textwidth]{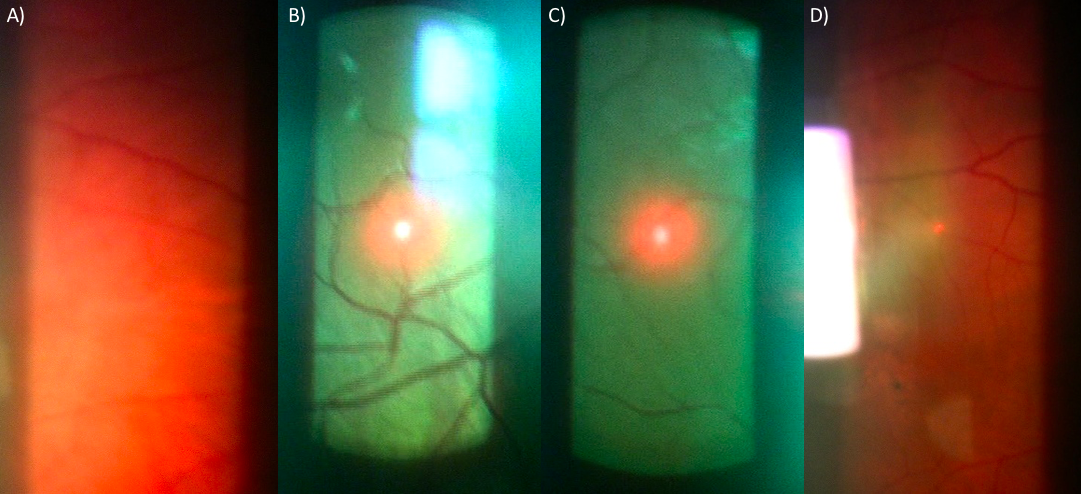}
\vspace{-2mm}\caption{Examples of images from the \textit{slit lamp} dataset showing challenging conditions for registration. From left to right: low vascularization and over-exposure leading to weak contrasts and lack of corners, motion blur, focus blur, acquisition artifacts and reflections.}
\label{fig:ex_dataset}
\vspace{-2mm}
\end{figure}

In this study we used the following test datasets:
\begin{enumerate}
\item The \textit{slit lamp} dataset: from retinal videos of 3 patients (different from the ones used for training), a random set of 206 frame pairs was selected as testing samples, with size \SIrange{338}{660}{px} by \SIrange{190}{350}{px}. Examples are shown in Figure~\ref{fig:ex_dataset}. The pairs were selected to have an overlap ranging from 20 to 100\%. They are related by affine transformations and rotations up to 15 degrees. 
Using a dedicated software tool, all pairs of images were manually annotated following common procedures~\cite{chen} with at least 5 corresponding points, which were then used to estimate the ground truth homographies relating the pairs. As the slit lamp images depict small area of retina, it is justified to apply the planar assumption in generating homographies~\cite{Cattin,giancardo}.

\item The \textit{\ac{FIRE}} dataset~\cite{Hernandez-matas2017}: a publicly available retinal image registration dataset with ground truth annotations. It consists of 129 retinal images forming 134 image pairs. The original images of 2912x2912 pixels were-down scaled to 15\% of their original size, to match the resolution of the training set. Examples of such images are shown in Figure~\ref{fig:matches}.
\end{enumerate}
As a pre-processing step for testing on fundus images, we isolated the green channel, applied adaptive histogram equalization and a bilateral filter to reduce noise and enhance the appearance of edges as proposed in~\cite{Zanet}. The effect of pre-processing can be seen in Figure~\ref{matches-slitlamp}.

Even though the focus of this paper is on the retinal images, we also tested the generalization capabilities of our model by evaluating it on natural images. We used the $Oxford$~\cite{oxford}, $EF$~\cite{Zitnik}, $Webcam$~\cite{Verdie2015, Jacobs2007} and $ViewPoint$~\cite{Yi2015} datasets. More details are provided in the supplementary material.

\subsection{Evaluation criteria}

We evaluated the performance using the following metrics:
\begin{enumerate}
    \item \textbf{Repeatability} describes the percentage of detected points $x \in P$ in image $I$ that are within an $\epsilon$-distance ($\epsilon=3$) to points $x' \in P'$ in $I'$ after transformation with $H_{I,I'}$, where $P$ and $P'$ are the sets of extracted points found in common regions to both images:

\begin{equation}
    \frac{\left | \left \{ x \in P, x' \in P'  \mid | \left \| H_{I,I'}*x - x' \right \| <\varepsilon\right \} \right |}{\left | P \right | + \left | P' \right |}
\end{equation}

    \item \textbf{Matching performance.} Matches were found using the \ac{NNDR} strategy, as proposed in~\cite{Lowe2004}: two keypoints are matched if the descriptor distance ratio between the first and the second nearest neighbor is below a certain threshold $t$.  Then, the following metrics were evaluated:
    
\begin{enumerate}[label=(\alph*)]
    \item \textbf{AUC}, area under the ROC curve created by varying the value of $t$, following ~\cite{Dahl2011,Winder,Windera}.  
    
    \item \textbf{M.score}, the ratio of correct matches over the total number of keypoints extracted by the detector in the shared viewpoint region~\cite{Mikolajczyk2005}.
    
    \item \textbf{Coverage fraction}, measures the coverage of an image by correctly matched key points. A coverage mask was generated from true positive key points, each one adding a disk of fixed radius (25px) as in~\cite{Aldana-iuit}. 
\end{enumerate}

We computed the homography $\hat{H}$ relating the reference to the transformed image  by applying \ac{RanSaC} algorithm to remove outliers from the detected matches.

\item \textbf{Registration success rate.} We furthermore evaluated the  registration accuracy achieved after using key points computed by different detectors as in~\cite{Chen2010,Wang2015}.  To do so, we compared the reprojection error of six fixed points of the reference image (denoted as $c_{i}, i=\left \{ 1,..,6 \right \}$) onto the other. For each image pair for which a homography was found, the quality of the registration was assessed with 
the median error (\textbf{MEE}) and the maximum error (\textbf{MAE}) of the distances between corresponding points after transformation.

Using these metrics, we defined different thresholds on MEE and MAE that define $"acceptable"$, $"inaccurate"$ and $"failed"$ registrations. We consider registration $"failed"$ if not enough keypoints or matches were found to compute a homography (minimum 4), if it involves a flip or if the estimated scaling component is greater than 4 or smaller than $0.1$. We classified the result as $"acceptable"$ when $MEE<10$ and $MAE<30$ and as $"inaccurate"$ otherwise. The values for the thresholds were found empirically by post-viewing the results. 
Using the above definitions, we calculated the \textbf{success rate of each class}, equal to the percentage of image pairs for which the registration falls into each category. These metrics are the most important quantitative evaluation criteria of the overall performance in a real-world setting.

\end{enumerate}

\subsection{Baselines and implementation details}

To evaluate the performance of our GLAMpoints detector associated with root-SIFT descriptor, we compared its matching ability and registration quality against well known detectors and descriptors. Among them, SIFT~\cite{sift}, KAZE~\cite{Alcantarilla2012} and LIFT~\cite{LIFT} were shown to perform well on fundus images by Truong \etal~\cite{truong}. Moreover, we compared our method to other CNN-based detectors-descriptors: \ac{LF-NET}~\cite{Trulls} and SuperPoint~\cite{Detone}. We used the authors' implementation of LIFT (pretrained on Picadilly), SuperPoint and \ac{LF-NET} (pretrained on indoor data, which gives better results on fundus images than the version pretrained on outdoor data) and OpenCV implementation for \ac{SIFT} and KAZE. A rotation-dependent version of root-\ac{SIFT} descriptor is used due to its better performance on our test set compared to the rotation invariant version. For the remainder of the paper, \ac{SIFT} descriptor refers to root-SIFT, rotation-dependent, except if otherwise stated.

Training of \ac{GLAMpoints} was performed using Tensorflow \cite{tensorflow2015-whitepaper} with mini-batch size of 5 and the Adam optimizer \cite{Kingma2015} with learning rate $= 0.001$ and $\beta$ = (0.9, 0.999) for 35 epochs. For each batch we randomly cropped $256\times 256$ patches of the full-resolution image to speed up the computation. \ac{GLAMpoints} (NMS10) was trained and tested with a \ac{NMS} window $w$ equal to 10px. It must be noted that other \ac{NMS} windows can be applied, which obtain similar performance.

\subsection{Quantitative results on the \textbf{\textit{slit lamp}} dataset}

\begin{figure*}
    \centering
    \includegraphics[width=0.99\textwidth]{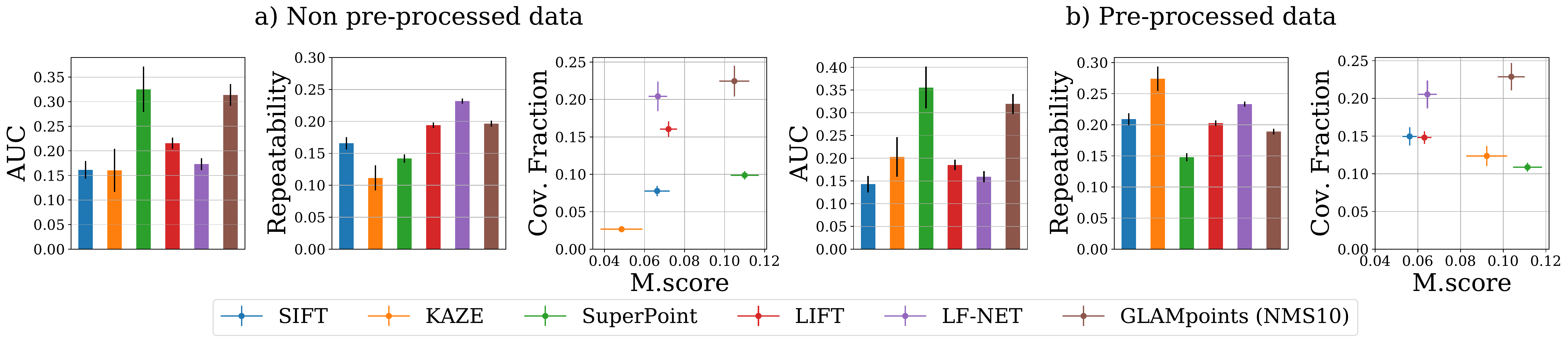}
    \caption{Summary of detector/descriptor performance metrics evaluated over 206 pairs of the \textit{slit lamp} dataset.}
    \label{results-slitlamp}
\end{figure*}

Table~\ref{tab:slitlamp_quantitative} presents the success rate of registration evaluated on the \textit{slit lamp} dataset. Without pre-processing, most detectors show lower performance compared to the pre-processed images, but GLAMpoints performs well even on raw images. While the success rate of acceptable registrations of \ac{SIFT}, KAZE and SuperPoint drops by 20 to 30\% between pre-processed and raw images, \ac{GLAMpoints} as well as \ac{LIFT} and \ac{LF-NET} show only a decrease of 3 to 6\%. Besides, \ac{LF-NET}, \ac{LIFT} and \ac{GLAMpoints} detect a steady average number of keypoints (around 485 for preprocessed and 350 non-preprocessed) independently of the pre-processing, whereas the other detectors see a reduction half. In general, GLAMpoints shows the highest performance for both raw and pre-processed images in terms of registration success rate. The robust results of our method indicate that while our detector performs as well or better on good quality images compared to the heuristic-based methods, its performance does not drop on lower quality images.

\begin{table}[t]
\centering
\caption{Success rates (\%) per registration class for each detector on the 206 images of the \textit{slit lamp} dataset. When the original descriptor is not used in association with detector, the descriptor used is indicated in parenthesis.}
(a) Raw data \\
\resizebox{0.44\textwidth}{!}{%
\begin{tabular}{llll}
\toprule
                  & Failed {[}\%{]} & Inaccurate {[}\%{]} & Acceptable {[}\%{]} \\ \midrule
SIFT        & 14.56  & 63.11                 & 22.33                 \\
KAZE        & 24.27  & 61.65    & 14.08                 \\
SuperPoint & 17.48  & 48.54                 & 33.98                 \\
LIFT        & 0.0    & 43.69                 & 56.31                 \\
\ac{LF-NET}    & 0.0                & 39.81                 & 60.19                 \\
GLAMpoints (SIFT)  & 0.0    & 36.41                 & \textbf{63.59}                 \\
\bottomrule
\end{tabular}%
}
\centering
\vspace{3mm}\\ (b) Pre-processed data \\
\resizebox{0.44\textwidth}{!}{%
\begin{tabular}{llll}
\toprule
Detector &  Failed {[}\%{]} & Inaccurate {[}\%{]} & Acceptable {[}\%{]} \\ \midrule
ORB  & 9.71 & 83.01 & 7.28 \\
GLAMpoints (ORB) & 0.0 & 88.35 & \textbf{11.65} \\
BRISK & 16.99 & 66.02 & 16.99 \\
GLAMpoints (BRISK) & 1.94 & 75.73 & \textbf{22.33}
\\ \midrule\midrule
SIFT & 1.94 & 47.75 & 50.49 \\
KAZE  & 1.46            & 54.85                          & 43.69  \\
KAZE (SIFT)  & 4.37               & 57.28                 & 38.35   \\
SuperPoint & 7.77            & 51.46                          & 40.78\\
SuperPoint (SIFT) & 6.80  & 54.37                 & 38.83  \\
LIFT & 0.0 & 39.81 & 60.19 \\
LF-NET & 0.0                & 36.89                 & 63.11       \\
LF-NET (SIFT) & 0.0                & 40.29                 & 59.71 \\
GLAMpoints (SIFT)  & 0.0             & 31.55                          & \textbf{68.45}
\\\midrule\midrule
Random grid (SIFT)  & 0.0    & 62.62                 & 37.38  \\
\bottomrule
\end{tabular}%
}
\label{tab:slitlamp_quantitative}\vspace{-3mm}
\end{table}

While \ac{SIFT} extracts a large number of keypoints (205.69 on average for unprocessed images and 431.03 for pre-processed), they appear in clusters as shown in Figure~\ref{matches-slitlamp}. As a result, even if the repeatability is relatively high, the close positioning of the interest points leads to a large number of rejected matches, as the nearest-neighbours are very close to each other. This is evidenced by the low coverage fraction, ${M.score}$ and ${AUC}$ (Figure~\ref{results-slitlamp}). With a similar value of repeatability, our approach extracts interest points widely spread and trained for their matching ability (highest \textit{coverage fraction}), resulting in more true positive matches (second highest \textit{M.score} and \textit{AUC}), as shown in Figure~\ref{results-slitlamp}. 

\ac{LF-NET}, similar to SIFT, shows high repeatability, which can be explained by its training strategy, which preferred repeatability over accurate matching objective. However, its \textit{M.score} and \textit{AUC} are in the bottom part of the ranking (Figure~\ref{results-slitlamp}). While the performance of \ac{LF-NET} may increase if it was trained on fundus images, its training procedure requires images pairs with their relative pose and corresponding depth maps, which would be extremely difficult - if not impossible - to obtain for fundus images. 

It is worth noting that SuperPoint scored the highest $M.score$ and $AUC$ but in this case the metrics are artificially inflated because very few keypoints are detected (35,88 and 59,21 on average for raw and pre-processed images respectively). This translates to  relatively small coverage fraction and one of the lowest repeatability, leading to few possible correct matches.

As part of an ablation study, we trained GLAMpoints with different descriptors (Table~\ref{tab:slitlamp_quantitative}b, top). While it performs best with the SIFT descriptor, the results show that for every considered descriptor (SIFT, ORB, BRISK), GLAMpoints improves upon the corresponding original detector.

To benchmark the detection results, we used the descriptors that were developed/trained jointly with the given detector and thus can be considered as optimal. For instance in~\cite{LIFT}, the combination of the LIFT/LIFT detector/descriptor outperformed LIFT/SIFT. For completeness, we present the registration results of baseline detectors combined with root-SIFT descriptor in Table~\ref{tab:slitlamp_quantitative}b, center.  As can be seen, using root-SIFT descriptor does not improve the result compared to the original descriptor.

Finally, to verify that the performance gain of GLAMpoints does not come solely from the uniform and dense spread of the detected keypoints, we computed the success rate for keypoints in a random, uniformly distributed grid (Table~\ref{tab:slitlamp_quantitative}b, bottom), which underperforms in comparison. This shows that our detector predicts not only uniform but also \textit{significant} points.

\begin{table}[t]
\centering
\caption{Success rates (\%) of each detector on \textit{FIRE}.}
\label{tab:success-rate-FIRE}
\vspace{-2mm}\resizebox{0.44\textwidth}{!}{%
\begin{tabular}{lllll}
\toprule
                                  & Failed {[}\%{]} & Inaccurate {[}\%{]} & Acceptable {[}\%{]} \\ \midrule
SIFT                              & 2.24 & 36.57                 & 61.19                 \\
KAZE                              & 14.18 & 58.21    & 27.61                 \\
SuperPoint                        & 0.0                & 13.43                 & 86.57     \\
LIFT                              & 0.0                & 10.45                 & 89.55           \\
LF-NET                            & 0.0 & 38.06 & 61.94 \\
GLAMpoints (OURS)                 & 0.0                & 5.22    & \textbf{94.78}        \\ \bottomrule
\end{tabular}%
}
\vspace{-4mm}
\label{tab:FIRE_threshold}
\end{table}

\subsection{Quantitative results on \textbf{\textit{FIRE}} dataset}

Table~\ref{tab:success-rate-FIRE} shows the results for success rates of registrations on FIRE. Our method outperforms baselines both in terms of success rate and global accuracy of non-failed registrations. As all the images in FIRE dataset present good quality with highly contrasted vascularization, we did not apply pre-processing. We also did not find it necessary to use the available background masks to filter out keypoints detected outside of the retina as generally they were not matched and did not contribute to the final registration. 

\begin{figure*}
\centering
\begin{tabular}{c c}
\includegraphics[width=0.37\textwidth]{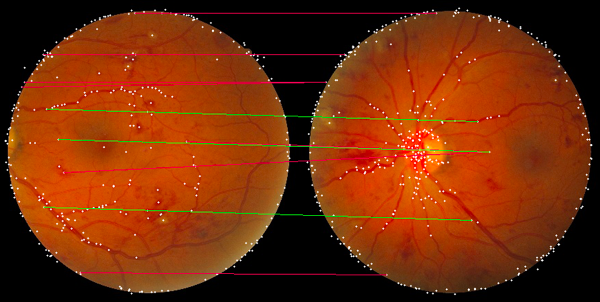} &
\includegraphics[width=0.37\textwidth]{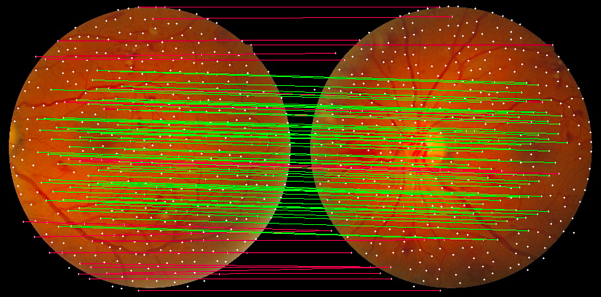} \\
\includegraphics[width=0.37\textwidth]{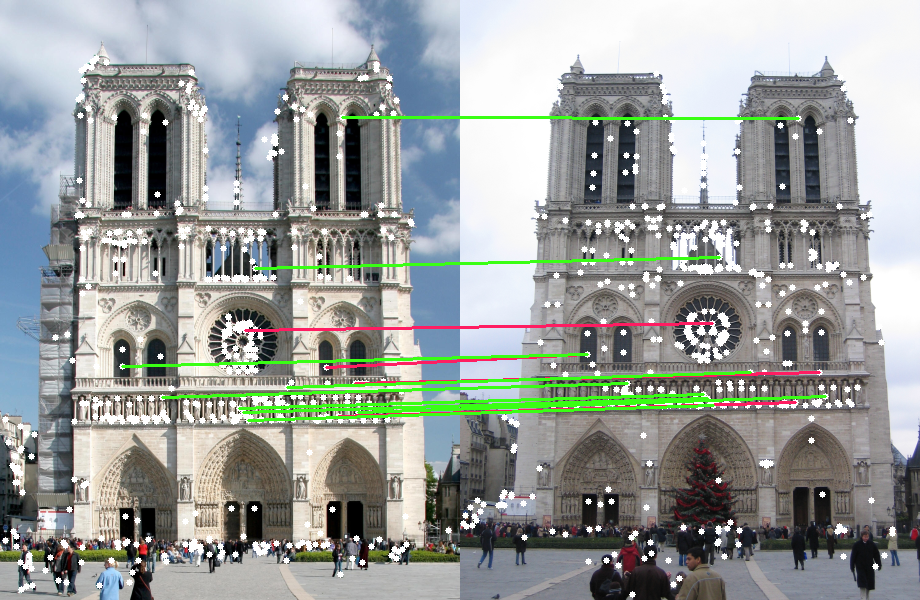} &
\includegraphics[width=0.37\textwidth]{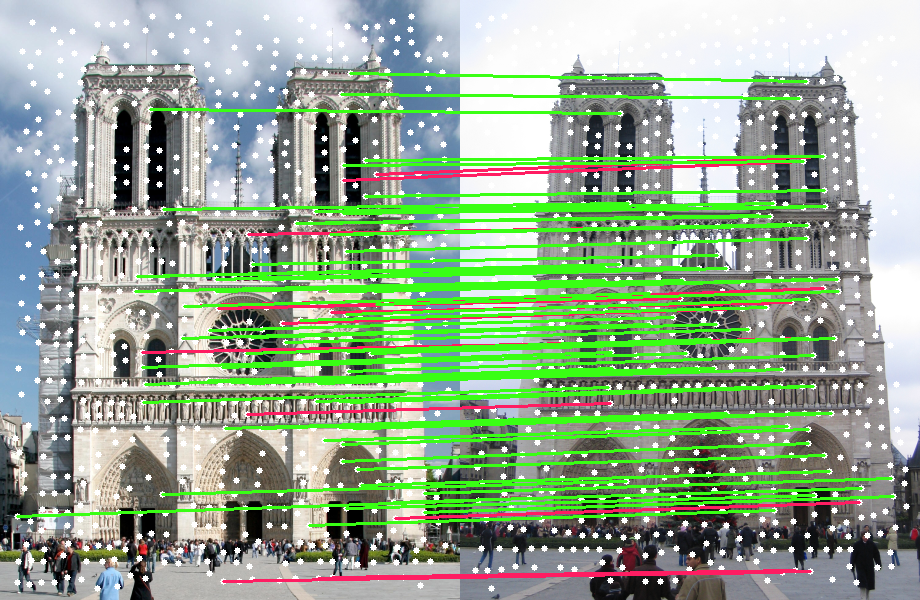} \\
a) SIFT & b) GLAMpoints\\
\end{tabular}
\vspace{-2mm}
\caption{Interest points detected by a) SIFT  and b) GLAMpoints and corresponding matches for a pair of images from the \textit{\ac{FIRE}} (top) and \textit{Oxford} (bottom) datasets. Detected points are in white, green matches correspond to true positive, red to false positive. GLAMpoints finds considerably more true positive points than SIFT. }
\label{fig:matches}
\vspace{-3mm}
\end{figure*}

It is interesting to note the gap of 33.6\% in the success rate of acceptable registrations between \ac{GLAMpoints} and \ac{SIFT}. As both use the same descriptor, this difference can be only explained by the quality of the detector. As can be seen in Figure~\ref{fig:matches}, \ac{SIFT} detects a restricted number of keypoints densely positioned solely on the vascular tree and in the image borders, while \ac{GLAMpoints} extracts interest points over the entire retina, including challenging areas such as the fovea and avascular zones, leading to a substantial rise in the number of correct matches. 

Even though \ac{GLAMpoints} outperforms all other detectors, \ac{LIFT} and SuperPoint also present high performance on the \textit{\ac{FIRE}} dataset. This dataset contains images with well-defined corners on a clearly contrasted vascular tree and \ac{LIFT} extracts keypoints spread over the entire image, while SuperPoint was trained to detect corners on synthetic primitive shapes. However, as evidenced on the \textit{slit lamp} dataset, the performance of SuperPoint strongly deteriorates on images with less clear features. 

\subsection{Results on natural images}

To further demonstrate a possible extension of our method to other image domains, we computed its predictions on natural images. Note that we used the same GLAMpoints model trained on slit lamp images. 

Globally, \ac{GLAMpoints} reaches a success rate of 75.38\% for acceptable registrations, against 85.13\% for the best performing detector - SIFT with rotation invariance - and 83.59\% for SuperPoint. In terms of $AUC$, $M.score$ and \textit{coverage fraction} it scores respectively second, second and first best. In contrast, $repeatability$ of \ac{GLAMpoints} is only second to last after \ac{SIFT}, KAZE and \ac{LF-NET} even though it successfully registers more images. This result shows once again that repeatability is not the most adequate metric to measure the performance of a detector. The detailed results can be found in the supplementary material. 

Finally, it should be noted that the outdoor images of this dataset are significantly different from medical fundus images and contain much greater variability of structures, which indicates a promising generalization of our model to unseen image domains.

\subsection{Qualitative results}

In case of slit lamp videos, the end goal is to create retinal mosaics. Using 10 videos containing 25 to 558 images, we generated mosaics by registering consecutive frames using keypoints detected by different methods. We calculated the average number of frames before the registration failed (due to the lack of extracted keypoints or correct matches between a pair of images). Over those 10 videos, the average number of registered frames before failure is 9.98 for \ac{GLAMpoints} and only 1.04 for \ac{SIFT}. 

\begin{figure}[t]
\centering
\begin{tabular}{c c c}
    \includegraphics[height=0.16\textwidth]{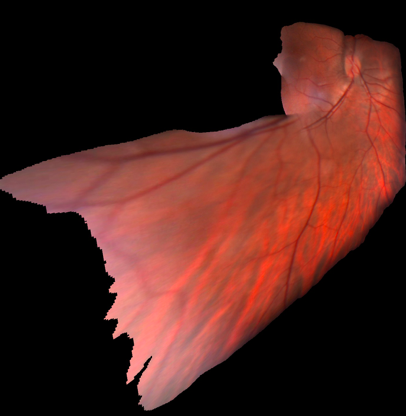} &
    \includegraphics[height=0.16\textwidth]{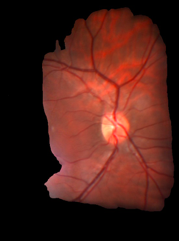} &
    \includegraphics[height=0.16\textwidth]{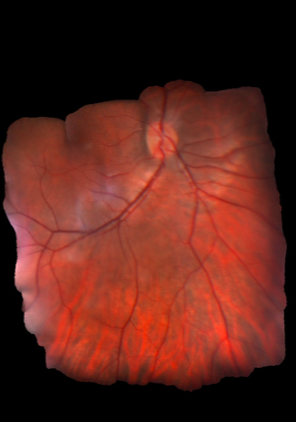} \\
    a) 11 frames & b)  34 frames & c) 53 frames\\
\end{tabular}
\vspace{-2mm}\caption{Mosaics obtained from registration of consecutive images until failure. a) SIFT, raw images; b) SIFT, pre-processed data; c) \ac{GLAMpoints}, raw data.}
\label{mosaics}
\end{figure}

Example mosaics are presented in Figure~\ref{mosaics}. For the same video, \ac{SIFT} failed after 34 frames when the data was pre-processed and only after 11 frames on the original data. In contrast, \ac{GLAMpoints} successfully registered 53 consecutive raw images, without visual errors. The mosaics were created with frame to frame matching with the blending method of~\cite{Zanet} and without bundle adjustment.

\subsection{Run time}

The run time of detection is computed over 84 pairs of images with a resolution of 660px by 350px. The \ac{GLAMpoints} architecture was run on a Nvidia GeForce GTX 1080 GPU while \ac{NMS} and \ac{SIFT} used CPU. Mean and standard deviation of run time for \ac{GLAMpoints} and \ac{SIFT} are presented in Table~\ref{run_time}. GLAMpoints is on average significantly faster than SIFT. Importantly, it does not require any time-consuming pre-processing step. 

\begin{table}
\centering
\caption{Average detection run time [ms] per image for GLAMpoints and \ac{SIFT} detectors.}
\vspace{-1mm}\resizebox{0.45\textwidth}{!}{%
\begin{tabular}{lll}
\toprule
                                    & \ac{GLAMpoints}         & SIFT              \\ \midrule
Pre-processing                      & 0.0                 & 16.64 $\pm$ 0.93  \\
Detection image I                   & \hspace{-2mm}\begin{tabular}{l} \textit{\ac{CNN}:} 16.28 $\pm$ 96.86 \\ 
\textit{NMS:} 11.2 $\pm$ 1.05\end{tabular} & 28.94 $\pm$ 1.88  \\

\textbf{Total}                  & \textbf{27.48} $\pm$ 98.74 & 45.58 $\pm$ 4.69  \\\bottomrule
\end{tabular}%
}\label{run_time}
\end{table}

%% file: 05_Conclusion.tex
\section{Conclusion}
\label{sec:conclusion}

In this paper we introduce GLAMpoints - a keypoint detector optimized for matching performance. This is in contrast to other detectors that are optimized for repeatability of keypoints, ignoring their correctness for matching. GLAMpoints detects significantly more keypoints that lead to correct matches even in low textured images, which do not present many features. As a result, no explicit pre-processing of the images is required. We train our detector on generated image pairs avoiding the need for ground truth correspondences. Our method produces state-of-the-art matching and registration results of medical fundus images and our experiments show that it can be further extended to other domains, such as natural images.

%% file: 06_Supplementary_Material.tex
\section*{Supplementary material} 

In this supplementary material, we first provide additional details on the training methodology in Section~\ref{training-sup}. We then give additional qualitative and quantitative evaluation results on fundus images in Section~\ref{fundus-sup}. Finally, in Section~\ref{natural-imges-sup}, we show the generalization capabilities of GLAMpoints by presenting evaluation results on natural images. Importantly, for all results, we use the same model weights trained on fundus images. For the entire supplementary material, \ac{SIFT} descriptor~\cite{Lowe2004} refers to root-SIFT~\cite{Arandjelovi}.

\section{Supplementary details on the training method}
\label{training-sup}

\subsection{Performance comparison between SIFT descriptor with or without rotation invariance}

\ac{GLAMpoints} detector was trained and tested in association with \ac{SIFT} descriptor rotation-dependent because \ac{SIFT} descriptor without rotation invariance performs better than the rotation invariant version on fundus images. The details of the metrics evaluated on the pre-processed \textit{slitlamp} dataset for both versions of \ac{SIFT} descriptor are shown in Table~\ref{tab:SIFT_metrics}.

\subsection{Method for homography generation}

For training of our GLAMpoints, we rely on pairs of images synthetically created by applying randomly sampled homography transformations to a set of base slitlamp images. 
Let B denote a particular base image from the training dataset, of size $H \times W$.
At every step $i$, an image pair $I_{i}, I'_{i}$ is generated from image $B$ by applying two separate, randomly sampled homography transforms $g_{i}, g'_{i}$. Each of those homographies is a composition of rotation, shearing, perspective, scaling and translation elements. The minimum and maximum values of the geometric transformation parameters that we used are given in table~\ref{tab:homo_generation}.

Nevertheless, it must be noted that the degree of the geometric transformations applied during training should be chosen so that the resulting synthetic training set resembles the test set. In our case, our test set composed of retinal images only showed rotation up to 30 degrees and only little scaling changes, therefore we limited the geometric transformations of the training set accordingly. However, any degree of geometric transformations or even non-linear ones can be applied to the original images to synthetically create pairs of training images.

\section{Details of results on fundus images}
\label{fundus-sup}

Here, we provide more detailed quantitative experiments on fundus images as well as additional qualitative results.

\subsection{Details of \textbf{\textit{MEE}} and \textbf{\textit{RMSE}} per registration class on the retinal images dataset}

Table~\ref{tab:details-registration-FIRE} and~\ref{tab:details-registration-slitlamp} show the mean and standard deviation of the median error $MEE$ and the root mean squared error $RMSE$ for respectively the \textit{FIRE} dataset and the $slitlamp$ dataset. In both cases, \ac{GLAMpoints} (NMS10) presents the highest registration accuracy for inaccurate registrations and globally.

\begin{table*}[t!]
\centering
\caption{Means and standard deviations of median errors (MEE) and RMSE in pixels for non-preprocessed images of the \textit{\ac{FIRE}} dataset. Acceptable registrations are defined as having ($MEE<10$ and $MAE <30$).}\label{tab:details-registration-FIRE}
\resizebox{\textwidth}{!}{%
\begin{tabular}{@{}lllllll@{}}
\toprule
                     & \multicolumn{2}{l}{\textbf{Inaccurate Registration}}                                 &        \multicolumn{2}{l}{\textbf{Acceptable Registration}}                                 &      
                       \multicolumn{2}{l}{\textbf{Global Non-Failed Registration}} \\
                     & \textit{MEE} & \textit{RMSE} & \textit{MEE} & \textit{RMSE} & \textit{MEE} & \textit{RMSE} \\
                      \midrule
SIFT                              & 66.44 $\pm$ 86.98   & 179.2 $\pm$ 412.88  & 3.79 $\pm$ 2.27    & 3.97 $\pm$ 2.3      & 27.22 $\pm$ 61.25  & 69.51 $\pm$ 266.37  \\
KAZE                              & 105.36 $\pm$ 118.94 & 314.0 $\pm$ 1184.76 & 4.69 $\pm$ 2.2     & 4.53 $\pm$ 2.28     & 72.97 $\pm$ 108.66 & 214.43 $\pm$ 986.38 \\
SuperPoint                        & 33.79 $\pm$ 69.65   & 48.97 $\pm$ 134.77  & 2.19 $\pm$ 2.12    & 2.2 $\pm$ 2.22      & 6.44 $\pm$ 27.78   & 8.48 $\pm$ 51.94    \\
LIFT                              & 24.39 $\pm$ 46.91   & 25.97 $\pm$ 43.93   & 2.4 $\pm$ 2.26     & 2.48 $\pm$ 2.54     & 4.7 $\pm$ 16.72    & 4.94 $\pm$ 16.09    \\
GLAMpoints (OURS)                 & \textbf{15.53 $\pm$ 7.89}    & \textbf{16.42 $\pm$ 6.26}    & \textbf{2.58 $\pm$ 2.36}    & \textbf{2.74 $\pm$ 2.54}     & \textbf{3.26 $\pm$ 4.1}     & \textbf{3.46 $\pm$ 4.17}     \\  \bottomrule
\end{tabular}%
}
\end{table*}

\begin{table*}[t!]
\centering
\caption{Means and standard deviations of median errors (MEE) and RMSE in pixels for the 206 images of the $slitlamp$ dataset. Acceptable registrations are defined as having ($MEE<10$ and $MAE <30$).}

(a) Raw data \\
\resizebox{\textwidth}{!}{%
\begin{tabular}{@{}lllllll@{}}
\toprule
                     & \multicolumn{2}{l}{\textbf{Inaccurate Registration}}                                 &        \multicolumn{2}{l}{\textbf{Acceptable Registration}}                                 &      
                       \multicolumn{2}{l}{\textbf{Global Non-Failed Registration}} \\
                     & \textit{MEE} & \textit{RMSE} & \textit{MEE} & \textit{RMSE} & \textit{MEE} & \textit{RMSE} \\
                      \midrule
SIFT        & 109.04 $\pm$ 132.13 & 368.56 $\pm$ 1766.18   & 5.15 $\pm$ 2.37 & 5.78 $\pm$ 2.45 & 81.89 $\pm$ 122.39 & 273.74 $\pm$ 1526.27 \\
KAZE        & 139.12 $\pm$ 123.07 & 640.8 $\pm$ 2980.66    & 5.58 $\pm$ 2.66 & 5.72 $\pm$ 2.39 & 114.29 $\pm$ 122.6 & 522.74 $\pm$ 2700.71 \\
SuperPoint & 131.82 $\pm$ 123.28 & 231.08 $\pm$ 509.82    & \textbf{3.82 $\pm$ 1.78} & \textbf{3.77 $\pm$ 1.71} & 79.12 $\pm$ 113.62 & 137.48 $\pm$ 406.7   \\
LIFT        & 114.25 $\pm$ 129.96 & 1335.03 $\pm$ 10820.78 & 3.94 $\pm$ 2.08 & 4.04 $\pm$ 2.04 & 52.14 $\pm$ 101.86 & 585.54 $\pm$ 7182.71 \\
LF-NET                  & 77.69 $\pm$ 112.34  & 92.97 $\pm$ 183.92     & 4.61 $\pm$ 2.28 & 4.62 $\pm$ 2.31 & 33.7 $\pm$ 79.41   & 39.79 $\pm$ 123.85   \\
GLAMpoints (OURS)  & \textbf{25.77 $\pm$ 38.32}   & \textbf{33.15 $\pm$ 85.49}      & 4.61 $\pm$ 2.16 & 4.6 $\pm$ 2.26  & \textbf{12.32 $\pm$ 25.32}  & \textbf{15.0 $\pm$ 53.41}     \\\bottomrule
\end{tabular}%
}

\centering
\vspace{3mm} (b) Pre-processed data \\
\resizebox{\textwidth}{!}{%
\begin{tabular}{@{}lllllll@{}}
\toprule
                     & \multicolumn{2}{l}{\textbf{Inaccurate Registration}}                                 &        \multicolumn{2}{l}{\textbf{Acceptable Registration}}                                 &      
                       \multicolumn{2}{l}{\textbf{Global Non-Failed Registration}} \\
                     & \textit{MEE} & \textit{RMSE} & \textit{MEE} & \textit{RMSE} & \textit{MEE} & \textit{RMSE} \\
                      \midrule
SIFT     & 65.2 $\pm$ 90.35    & 130.55 $\pm$ 273.75  & 4.92 $\pm$ 2.15    & 5.01 $\pm$ 2.25     & 34.17 $\pm$ 69.79  & 65.92 $\pm$ 200.74   \\
KAZE              & 86.83 $\pm$ 117.22  & 870.24 $\pm$ 7016.58 & 4.33 $\pm$ 2.26    & 4.45 $\pm$ 2.43     & 50.26 $\pm$ 96.6   & 486.39 $\pm$ 5252.64 \\
SuperPoint       & 117.53 $\pm$ 125.01 & 194.5 $\pm$ 312.9    & 4.21 $\pm$ 2.03    & 4.11 $\pm$ 2.05     & 67.43 $\pm$ 109.03 & 110.33 $\pm$ 252.12  \\
LIFT              & 113.3 $\pm$ 134.58  & 1328.06 $\pm$ 8854.49 & \textbf{4.15 $\pm$ 2.25}    & \textbf{4.21 $\pm$ 2.36}     & 47.6 $\pm$ 100.34 & 531.18 $\pm$ 5623.92 \\
LF-NET                   & 75.34 $\pm$ 128.6   & 158.78 $\pm$ 473.55   & 4.41 $\pm$ 2.16 & 4.45 $\pm$ 2.23 & 30.58 $\pm$ 85.3   & 61.39 $\pm$ 297.12   \\
GLAMpoints (OURS)       & \textbf{30.13 $\pm$ 56.86}   & \textbf{27.53 $\pm$ 42.41}    & 4.85 $\pm$ 2.44    & 4.85 $\pm$ 2.47     & \textbf{12.83 $\pm$ 34.09}  & \textbf{12.01 $\pm$ 26.13}    \\
 \bottomrule
\end{tabular}%
}\label{tab:details-registration-slitlamp}
\end{table*}

\subsection{Supplementary examples of matching on the \textbf{\textit{FIRE}} dataset}

\begin{figure*}
\centering
\vspace{-3mm}\includegraphics[width=0.86\textwidth]{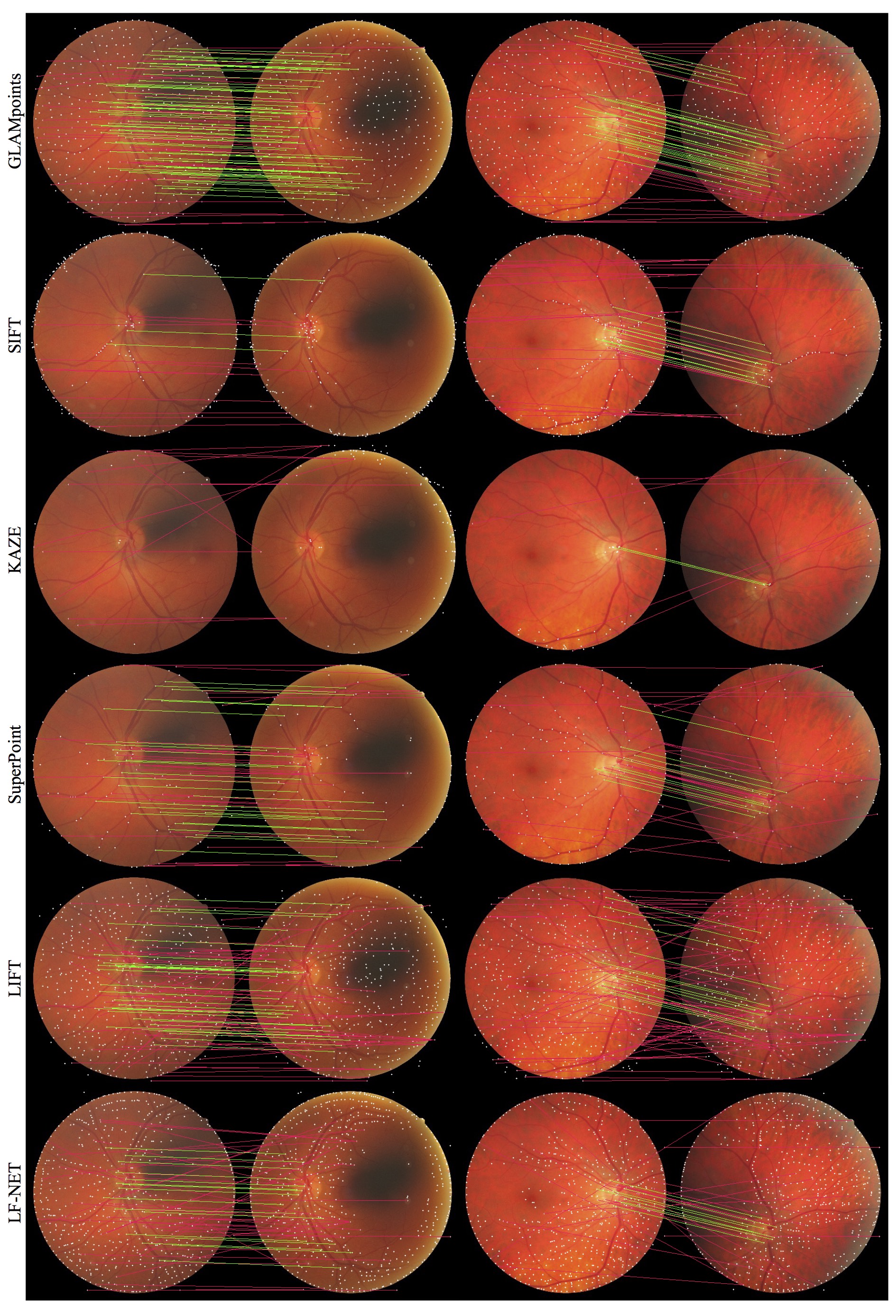}
\vspace{-5mm}\caption{Matches on the $FIRE$ dataset. Detected points are in white, green lines are true positive matches while red ones are false positive.}\label{fig:sup_matches}
\end{figure*}

We show additional examples of matches obtained by \ac{GLAMpoints}, \ac{SIFT}, KAZE, SuperPoint, \ac{LIFT} and \ac{LF-NET} on two pairs of
images from the $FIRE$ dataset in Figure~\ref{fig:sup_matches}. Again, our keypoints GLAMpoints are homogeneously spread and they lead to substantially more true-positive matches (in green in the figure) than any other method.

\section{Generalization of the model on natural images}
\label{natural-imges-sup}

Our method GLAMpoints was designed for application on medical retinal images. However, to show its generalisation properties, we also evaluate our network on natural images. Importantly, it must be noted that here, we use the model weights trained on slitlamp images. 

\begin{figure*}[t]
\centering
\includegraphics[width=0.99\textwidth]{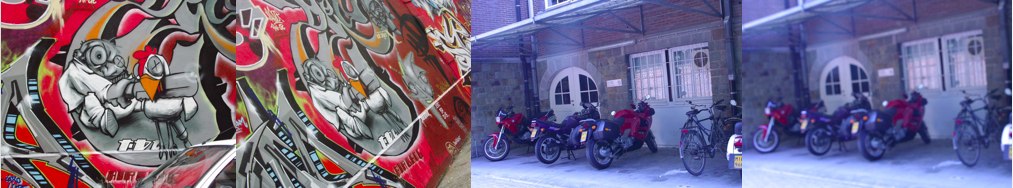}
\caption{Examples of image pairs from the Oxford dataset. }\label{fig:ex-oxford}
\end{figure*}

Our method was tested on several natural image datasets, with following specifications: 
\begin{enumerate}
\item $Oxford$ dataset~\cite{oxford}: 8  sequences  with  45  pairs  in total.   The  dataset  contains  various  imaging  changes including viewpoint, rotation, blur, illumination, scale, JPEG compression changes. We evaluated on six of these sequences, excluding the ones showing rotation (\textit{boat} and \textit{bark}). Indeed, we trained our model associated with SIFT descriptor without rotation invariance. To be consistent, SIFT descriptor rotation-dependent was also used for testing. 
\item $ViewPoint$ dataset~\cite{Yi2015}: 5 sequences with 25 pairs in total. It exhibits large viewpoint changes and in-plane rotations up to 45 degrees.
\item $EF$ dataset~\cite{Zitnik}:  3 sequences with 17 pairs in total. The dataset exhibits drastic lighting changes as well as daytime changes and viewpoint changes.
\item $Webcam$ dataset~\cite{Verdie2015, Jacobs2007}: 6 sequences with 124 pairs in total. It shows seasonal changes as well as day time changes of scenes taken from far away.
\end{enumerate}

For all of the aforementioned datasets, the images pairs are related by homography transforms. Indeed, the scenes are either planar, purely rotative or the images are taken at sufficient distance so that the planar assumption holds. In figure~\ref{fig:ex-oxford} are represented examples of images pairs from the Oxford dataset.

\begin{figure*}
\centering
    \includegraphics[width=0.99\textwidth]{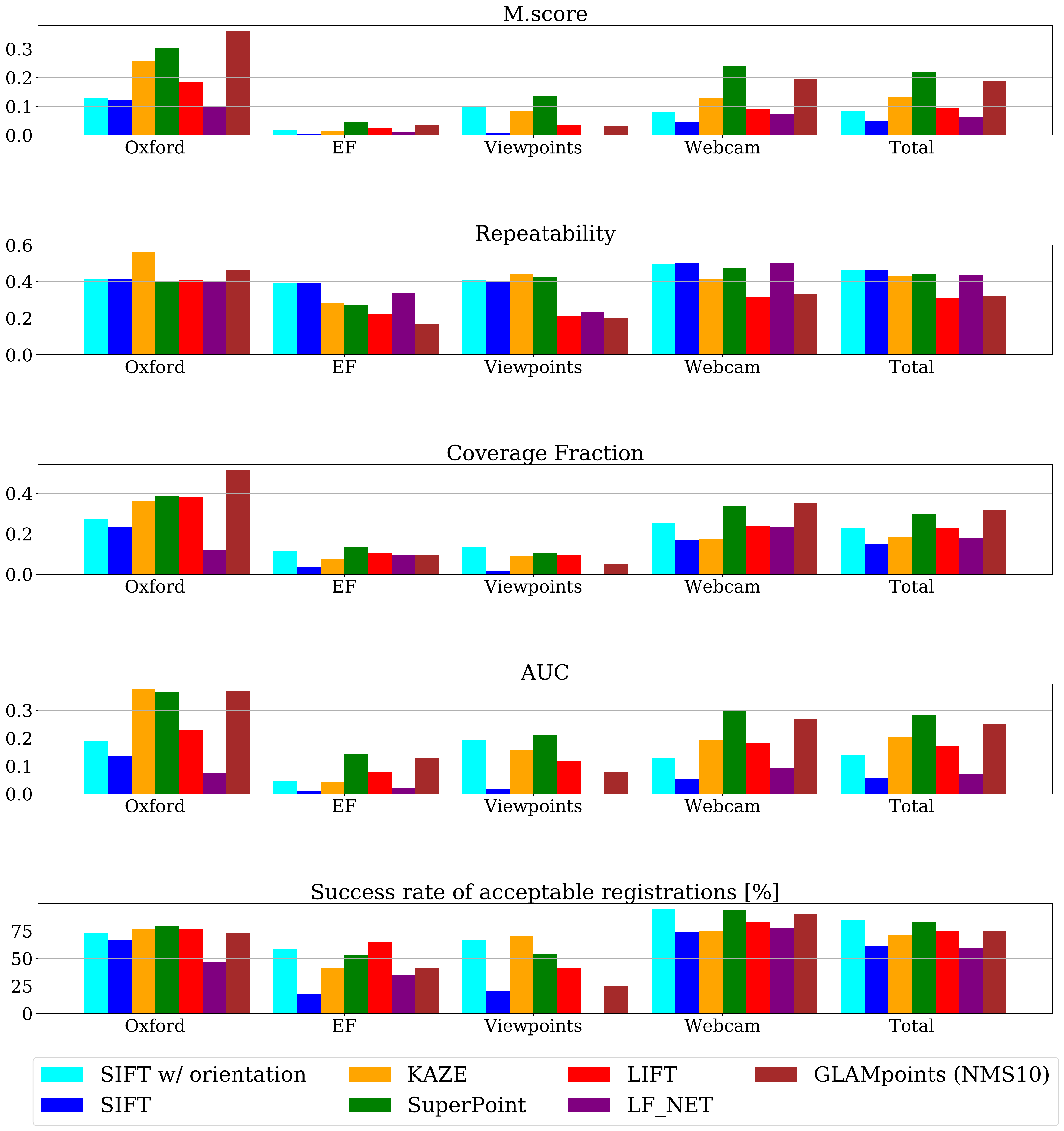}
    \caption{Summary of detector/descriptor performance metrics evaluated over 195 pairs of natural images.}\label{fig:results-oxford}
\end{figure*}

The metrics computed on the aforementioned datasets are shown in Figure~\ref{fig:results-oxford}. We use the same thresholds as in the main paper to determine acceptable, inaccurate and failed registration. We used the LF-NET pretrained on outdoor data, since most images of those datasets are outdoor. It is worth mentioning the gap in performance between \ac{SIFT} descriptor with or without rotation invariance on the $EF$ and the $Viewpoints$ datasets. Those images exhibit large rotations and therefore a rotation invariant descriptor is necessary, which is not currently the case of our detector associated with \ac{SIFT}. This explains why \ac{GLAMpoints} performs poorly on those datasets. 

Besides, it is interesting to note that on the \textit{Viewpoints} dataset, \ac{LF-NET} scores extremely low in all metrics except for $repeatability$. Indeed, on those images, even though the extracted key-points are repeatable, most of them are useless for matching. Therefore, LF-Net finds only very few true positive matches compared to the number of detected keypoints and matches, leading to poor evaluation results. This emphasize the importance of designing a detector specifically optimized for matching purposes. 

Finally, on the $Oxford$ dataset, \ac{GLAMpoints} outperforms all other detectors in terms of $M.score$, \textit{coverage fraction} and $AUC$ while scoring second in repeatability. 

\newpage

Finally, to further extend GLAMpoints to natural images, which often show more drastic view-point changes than retinal images, one could adapt the parameters of the synthetic homographies and appearance augmentations applied during training to fit better the test set of interest. Indeed, in the case of slitlamp retinal images, our test set composed of retinal images only showed rotation up to 30 degrees and only little scaling changes, therefore we limited the geometric transformations of the training set accordingly. However, any degree of geometric transformation or even non-linear ones can be applied to the original images to synthetically create pairs of training images.

%% file: ms.bbl
\begin{thebibliography}{10}\itemsep=-1pt

\bibitem{brute}
{OpenCV: cv::BFMatcher Class Reference}.

\bibitem{sift}
{OpenCV: cv::xfeatures2d::SIFT Class Reference}.

\bibitem{tensorflow2015-whitepaper}
Martín Abadi, Ashish Agarwal, Paul Barham, Eugene Brevdo, Zhifeng Chen, Craig
  Citro, Greg Corrado, Andy Davis, Jeffrey Dean, Matthieu Devin, Sanjay
  Ghemawat, Ian Goodfellow, Andrew Harp, Geoffrey Irving, Michael Isard,
  Yangqing Jia, Rafal Jozefowicz, Lukasz Kaiser, Manjunath Kudlur, Josh
  Levenberg, Dan Mané, Rajat Monga, Sherry Moore, Derek Murray, Chris Olah,
  Mike Schuster, Jonathon Shlens, Benoit Steiner, Ilya Sutskever, Kunal Talwar,
  Paul Tucker, Vincent Vanhoucke, Vijay Vasudevan, Fernanda Viégas, Oriol
  Vinyals, Pete Warden, Martin Wattenberg, Martin Wicke, Yuan Yu, and Xiaoqiang
  Zheng.
\newblock { {TensorFlow}: Large-Scale Machine Learning on Heterogeneous
  Systems}, 2015.

\bibitem{Alahi2012}
Alexandre Alahi, Raphaël Ortiz, and Pierre Vandergheynst.
\newblock {FREAK: Fast Retina Keypoint}.
\newblock In {\em Conference on Computer Vision and Pattern Recognition}, 2012.

\bibitem{Alcantarilla2012}
Pablo~Fern\'{a}ndez Alcantarilla, Adrien Bartoli, and Andrew~J. Davison.
\newblock {KAZE Features}.
\newblock In {\em European Conference on Computer Vision}, 2012.

\bibitem{Aldana-iuit}
Javier Aldana{-}Iuit, Dmytro Mishkin, Ondrej Chum, and Jiri Matas.
\newblock {In the Saddle: Chasing Fast and Repeatable Features}.
\newblock In {\em International Conference on Pattern Recognition}, pages
  675--680, 2016.

\bibitem{Altwaijry16}
Hani Altwaijry, Eduard Trulls, Serge Belongie, James Hays, and Pascal Fua.
\newblock {Learning to Match Aerial Images with Deep Attentive Architecture}.
\newblock In {\em Conference on Computer Vision and Pattern Recognition}, 2016.

\bibitem{Altwaijry2016b}
Hani Altwaijry, Andreas Veit, and Serge Belongie.
\newblock {Learning to Detect and Match Keypoints with Deep Architectures}.
\newblock In {\em British Machine Vision Conference}, 2016.

\bibitem{Arandjelovi}
Relja Arandjelovic and Andrew Zisserman.
\newblock {Three things everyone should know to improve object retrieval}.
\newblock In {\em Conference on Computer Vision and Pattern Recognition}, pages
  2911--2918, 2012.

\bibitem{Johns}
Vassileios Balntas, Edward Johns, Lilian Tang, and Krystian Mikolajczyk.
\newblock {PN-Net: Conjoined Triple Deep Network for Learning Local Image
  Descriptors}.
\newblock {\em CoRR}, abs/1601.05030, 2016.

\bibitem{Lenc}
Vassileios Balntas, Karel Lenc, Andrea Vedaldi, and Krystian Mikolajczyk.
\newblock {HPatches: A Benchmark and Evaluation of Handcrafted and Learned
  Local Descriptors}.
\newblock In {\em Conference on Computer Vision and Pattern Recognition}, pages
  3852--3861, 2017.

\bibitem{Bay2006}
Herbert Bay, Tinne Tuytelaars, and Luc Van~Gool.
\newblock {Surf: Speeded up robust features}.
\newblock In {\em European Conference on Computer Vision}, pages 404--417,
  2006.

\bibitem{Cattin}
Philippe~C. Cattin, Herbert Bay, Luc Van~Gool, and G{\'a}bor Sz{\'e}kely.
\newblock {Retina Mosaicing Using Local Features}.
\newblock In {\em Conference on Medical Image Computing and Computer Assisted
  Intervention}, pages 185--192, 2006.

\bibitem{chen}
J. {Chen}, J. {Tian}, N. {Lee}, J. {Zheng}, R.~T. {Smith}, and A.~F. {Laine}.
\newblock {A Partial Intensity Invariant Feature Descriptor for Multimodal
  Retinal Image Registration}.
\newblock {\em IEEE Transactions on Biomedical Engineering}, 57(7):1707--1718,
  2010.

\bibitem{Chen2010}
Jian Chen, Jie Tian, Noah Lee, Jian Zheng, Theodore~R. Smith, and Andrew~F.
  Laine.
\newblock {A Partial Intensity Invariant Feature Descriptor for Multimodal
  Retinal Image Registration}.
\newblock {\em IEEE Transactions on Biomedical Engineering}, 57(7):1707--1718,
  2010.

\bibitem{Cideciyan1995}
Artur~V. Cideciyan.
\newblock {Registration of Ocular Fundus Images: an Algorithm Using
  Cross-correlation of Triple Invariant Image Descriptors}.
\newblock {\em {IEEE} Engineering in Medicine and Biology Magazine},
  14(1):52--58, 1995.

\bibitem{Dahl2011}
Anders~L. Dahl, Henrik Aan{\ae}s, and Kim~S. Pedersen.
\newblock {Finding the Best Feature Detector-Descriptor Combination}.
\newblock In {\em International Conference on 3D Imaging, Modeling, Processing,
  Visualization and Transmission}, pages 318--325, 2011.

\bibitem{slam}
Andrew Davison.
\newblock {Real-Time Simultaneous Localisation and Mapping with a Single
  Camera}.
\newblock In {\em International Conference on Computer Vision}, 2003.

\bibitem{Zanet}
Sandro {De Zanet}, Tobias Rudolph, Rogerio Richa, Christoph Tappeiner, and
  Raphael Sznitman.
\newblock {Retinal Slit Lamp Video Mosaicking}.
\newblock {\em International Journal of Computer Assisted Radiology and
  Surgery}, 11(6):1035--1041, 2016.

\bibitem{Detone}
Daniel DeTone, Tomasz Malisiewicz, and Andrew Rabinovich.
\newblock {SuperPoint: Self-Supervised Interest Point Detection and
  Description}.
\newblock In {\em Conference on Computer Vision and Pattern Recognition
  Workshops}, pages 224--236, 2018.

\bibitem{Febbo18}
Paolo Di~Febbo, Carlo Dal~Mutto, Kinh Tieu, and Stefano Mattoccia.
\newblock {KCNN: Extremely-Efficient Hardware Keypoint Detection With a Compact
  Convolutional Neural Network}.
\newblock In {\em Conference on Computer Vision and Pattern Recognition
  Workshops}, 2018.

\bibitem{Fisher}
Philipp Fischer, Alexey Dosovitskiy, and Thomas Brox.
\newblock {Descriptor Matching with Convolutional Neural Networks: a Comparison
  to SIFT }.
\newblock Technical Report 1405.5769, arXiv, May 2014.

\bibitem{RANSAC}
Martin~A. Fischler and Robert~C. Bolles.
\newblock {Random sample consensus: a paradigm for model fitting with
  applications to image analysis and automated cartography}.
\newblock {\em Communications of the ACM}, 24(6):381--395, June 1981.

\bibitem{giancardo}
Luca Giancardo, Fabrice Meriaudeau, Thomas Karnowski, Tobin Kenneth~W., Jr,
  Enrico Grisan, Paolo Favaro, Alfredo Ruggeri, and Edward Chaum.
\newblock {Textureless Macula Swelling Detection With Multiple Retinal Fundus
  Images}.
\newblock {\em IEEE Transactions on Biomedical Engineering}, 58(3):795--799,
  2011.

\bibitem{Hang}
Yiliu Hang, Xiaofeng Zhang, Yeqin Shao, Huiqun Wu, and Wei Sun.
\newblock {Retinal Image Registration Based on the Feature of Bifurcation
  Point}.
\newblock In {\em International Congress on Image and Signal Processing,
  BioMedical Engineering and Informatics}, 2017.

\bibitem{harris}
Chris Harris and Mike Stephens.
\newblock {A Combined Corner and Edge Detector}.
\newblock In {\em Fourth Alvey Vision Conference}, 1988.

\bibitem{Hernandez-matas2017}
Carlos Hernandez-Matas, Xenophon Zabulis, Areti Triantafyllou, Panagiota
  Anyfanti, Stella Douma, and Antonis Argyros.
\newblock {FIRE : Fundus Image Registration dataset}.
\newblock {\em Journal for Modeling in Ophthalmology}, 4:16--28, 2017.

\bibitem{Jacobs2007}
Nathan Jacobs, Nathaniel Roman, and Robert Pless.
\newblock {Consistent Temporal Variations in Many Outdoor Scenes}.
\newblock In {\em Conference on Computer Vision and Pattern Recognition}, 2007.

\bibitem{Huang2005}
Li~Ma Jun-Zhou~Huang, Tie-Niu~Tan and Yun-Hong Wang.
\newblock {Phase Correlation-based Iris Image Registration Model}.
\newblock {\em Journal of Computer Science and Technology}, 20(3):419--425,
  2005.

\bibitem{Kingma2015}
Diederik.~P. Kingma and Jimmy Ba.
\newblock {Adam: {A} Method for Stochastic Optimisation}.
\newblock In {\em International Conference on Learning Representations}, 2015.

\bibitem{Legg2013}
Phil Legg, Paul Rosin, David Marshall, and James Morgan.
\newblock {Improving Accuracy and Efficiency of Mutual Information for
  Multi-modal Retinal Image Registration using Adaptive Probability Density
  Estimation}.
\newblock {\em Computerized Medical Imaging and Graphics}, 37(7-8):597--606,
  2013.

\bibitem{Leutenegger2011}
Stefan Leutenegger, Margarita Chli, and Roland Siegwart.
\newblock {{BRISK}: Binary Robust Invariant Scalable Keypoints}.
\newblock In {\em International Conference on Computer Vision}, 2011.

\bibitem{Li2017}
P. Li, Q. Chen, W. Fan, and S. Yuan.
\newblock {Registration of {OCT} Fundus Images with Color Fundus Images Based
  on Invariant Features}.
\newblock In {\em Cloud Computing and Security}, pages 471--482, 2017.

\bibitem{Lowe2004}
David~G. Lowe.
\newblock {Distinctive Image Features from Scale-Invariant Keypoints}.
\newblock {\em International Journal of Computer Vision}, 20(2):91--110, Nov
  2004.

\bibitem{Mikolajczyk2005}
Krystian Mikolajczyk and Cordelia Schmid.
\newblock {A Performance Evaluation of Local Descriptors}.
\newblock {\em IEEE Transactions on Pattern Analysis and Machine Intelligence},
  27(10):1615--1630, 2005.

\bibitem{oxford}
Krystian Mikolajczyk, Tinne Tuytelaars, Cordelia Schmid, Andrew Zisserman, Jiri
  Matas, Frederik Schaffalitzky, Timor Kadir, and Luc {Van~Gool}.
\newblock {A Comparison of Affine Region Detectors}.
\newblock {\em International Journal of Computer Vision}, 65(1/2):43--72, 2005.

\bibitem{Trulls}
Yuki Ono, Eduard Trulls, Pascal Fua, and Kwang~Moo Yi.
\newblock {LF-Net: Learning Local Features from Images}.
\newblock In {\em Advances in Neural Information Processing Systems}, pages
  6237--6247, 2018.

\bibitem{Pluim2003}
Josien P.~W. Pluim, J.~B.~Antoine Maintz, and Max~A. Viergever.
\newblock {Mutual Information Based Registration of Medical Images: {A}
  Survey}.
\newblock {\em IEEE Transactions on Medical Imaging}, 22(8):986--1004, 2003.

\bibitem{Ramli}
Roziana Ramli, Mohd Yamani~Idna Idris, Khairunnisa Hasikin, Noor~Khairiah
  A~Karim, Ainuddin~Wahid Abdul~Wahab, Ismail Ahmedy, Fatimah Ahmedy,
  Nahrizul~Adib Kadri, and Hamzah Arof.
\newblock {Feature-Based Retinal Image Registration Using D-Saddle Feature}.
\newblock {\em Journal of Healthcare Engineering}, 2017:1--15, 10 2017.

\bibitem{ronneberger}
Olaf Ronneberger, Philipp Fischer, and Thomas Brox.
\newblock {U-Net: Convolutional Networks for Biomedical Image Segmentation}.
\newblock In {\em Conference on Medical Image Computing and Computer Assisted
  Intervention}, pages 234--241, 2015.

\bibitem{Rublee2011}
Ethan Rublee, Vincent Rabaud, Kurt Konolige, and Gary Bradski.
\newblock {ORB: An Efficient Alternative to SIFT or SURF}.
\newblock In {\em International Conference on Computer Vision}, 2011.

\bibitem{Sanchez-galeana2001}
C{\'e}sar~A S{\'a}nchez-Galeana, Christopher Bowd, Eytan~Z. Blumenthal,
  Parag~A. Gokhale, Linda~M. Zangwill, and Robert~N. Weinreb.
\newblock {Using Optical Imaging Summary Data to Detect Glaucoma}.
\newblock {\em Opthamology}, pages 1812--1818, 2001.

\bibitem{Simo-serra15}
Edgar {Simo-Serra}, Eduard Trulls, Luis Ferraz, Iasonas Kokkinos, Pascal Fua,
  and Franscesc {Moreno-Noguer}.
\newblock {Discriminative Learning of Deep Convolutional Feature Point
  Descriptors}.
\newblock In {\em International Conference on Computer Vision}, 2015.

\bibitem{bundle}
Bill Triggs, Philip~F. McLauchlan, Richard~I. Hartley, and Andrew~W.
  Fitzgibbon.
\newblock {Bundle Adjustment -- A Modern Synthesis}.
\newblock In {\em Vision Algorithms: Theory and Practice}, pages 298--372,
  2000.

\bibitem{glampoint_pytorch}
Prune Truong.
\newblock {GLAMpoints : Github project page in PyTorch.}
\newblock \url{https://github.com/PruneTruong/GLAMpoints_pytorch}, 2019.

\bibitem{glampoint_tensorflow}
Prune Truong, Stefanos Apostolopoulos, Agata Mosinska, Samuel Stucky, Carlos
  Ciller, and Sandro~De Zanet.
\newblock {GLAMpoints : GitLab project page in TensorFlow.}
\newblock \url{https://gitlab.com/retinai_sandro/glampoints}, 2019.

\bibitem{truong}
Prune Truong, Sandro {De Zanet}, and Stefanos Apostolopoulos.
\newblock {Comparison of Feature Detectors for Retinal Image Alignment}.
\newblock In {\em ARVO}, 2019.

\bibitem{Verdie2015}
Yannick Verdie, Kwang~Moo Yi, Pascal Fua, and Vincent Lepetit.
\newblock {TILDE: A Temporally Invariant Learned DEtector}.
\newblock {\em Conference on Computer Vision and Pattern Recognition}, pages
  5279--5288, 2015.

\bibitem{Wang2015}
Gang Wang, Zhicheng Wang, Yufei Chen, and Weidong Zhao.
\newblock {Robust Point Matching Method for Multimodal Retinal Image
  Registration}.
\newblock {\em Biomedical Signal Processing and Control}, 19:68--76, 2015.

\bibitem{Windera}
Simon Winder and Matthew Brown.
\newblock {Learning Local Image Descriptors}.
\newblock In {\em Conference on Computer Vision and Pattern Recognition}, June
  2007.

\bibitem{Winder}
Simon Winder, Gang Hua, and Matthew. Brown.
\newblock {Picking the Best {DAISY}}.
\newblock In {\em Conference on Computer Vision and Pattern Recognition}, pages
  178--185, 2009.

\bibitem{LIFT}
Kwang~Moo Yi, Eduard Trulls, Vincent Lepetit, and Pascal Fua.
\newblock {{LIFT:} Learned Invariant Feature Transform}.
\newblock In {\em European Conference on Computer Vision}, pages 467--483,
  2016.

\bibitem{Yi2015}
Kwang~Moo Yi, Yannick Verdie, Pascal Fua, and Vincent Lepetit.
\newblock {Learning to Assign Orientations to Feature Points}.
\newblock In {\em Conference on Computer Vision and Pattern Recognition}, 2016.

\bibitem{Zhou1994}
Liang Zhou, Mark~S. Rzeszotarski, Lawrence~J. Singerman, and Jeanne~M.
  Chokreff.
\newblock {The Detection and Quantification of Retinopathy Using Digital
  Angiograms}.
\newblock {\em IEEE Transactions on Medical Imaging}, 13(4):619--626, 1994.

\bibitem{Zitnik}
Larry Zitnick and Krishnan Ramnath.
\newblock {Edge Foci Interest Points}.
\newblock In {\em International Conference on Computer Vision}, 2011.

\end{thebibliography}
